\documentclass{article}

 \usepackage[final]{nips_2018}

\usepackage[utf8]{inputenc} 
\usepackage[T1]{fontenc}    
\usepackage{hyperref}       
\usepackage{url}    
\usepackage{cite}        
\usepackage{booktabs}       
\usepackage{amsfonts}       
\usepackage{nicefrac}       
\usepackage{amsmath}
\usepackage{graphicx}
\usepackage{subfigure}
\usepackage{color}
\usepackage{multirow}
\usepackage{soul}

\title{Leveraging Deep Stein's Unbiased Risk Estimator for Unsupervised X-ray Denoising}

\author{
  Fahad~Shamshad$^{1}$, Muhammad Awais$^{1}$, Muhammad Asim$^{1}$, \\ \textbf{Zain ul Aabidin Lodhi$^{1}$, Muhammad Umair$^{2}$, Ali Ahmed$^{1}$}  \\
  $^{1}$Department of Electrical Engineering, Information Technology University, Lahore, Pakistan \\
  $^{2}$Department of Civil Engineering, University of Toronto, Canada \\
}

\begin{document}

\maketitle

\begin{abstract}
Among the plethora of techniques devised to curb the prevalence of noise in medical images, deep learning based approaches have shown the most promise. However, one critical limitation of these deep learning based denoisers is the requirement of high-quality noiseless ground truth images that are difficult to obtain in many medical imaging applications such as X-rays. To circumvent this issue, we leverage recently proposed approach of \citep{soltanayev2018training} that incorporates Stein's Unbiased Risk Estimator (SURE) to train a deep convolutional neural network without requiring denoised ground truth X-ray data. Our experimental results demonstrate the effectiveness of SURE based approach for denoising X-ray images.


\end{abstract}
\section{Introduction}

X-ray images provide crucial support for diagnosis and decision making in many diverse clinical applications.  However, X-ray images may be corrupted by statistical noise, thus seriously deteriorating the quality and raising the difficulty of diagnosis \citep{thanhreview}. Therefore, X-ray denoising is an essential pre-processing step for improving the quality of raw X-ray images and their relevant clinical information content. 

Deep learning with massive amounts of training data has revolutionized many image processing and computer vision tasks including image denoising \citep{lecun2015deep}. Deep learning based denoisers have been recently shown to produce state of the art results \citep{zhang2017beyond}, and have been extensively investigated for denoising X-ray images for enhanced diagnosis reliability \citep{gondara2016medical, chen2017low}. These deep learning based denoisers are usually trained by minimizing mean squared error (MSE). This requires access to abundant high quality and clean ground truth X-ray images that are hard to acquire. 

In this work, we leverage recently proposed approach of \citep{soltanayev2018training} to train a deep convolutional neural network for denoising, using only noisy X-ray data. Denoising approach of \citep{soltanayev2018training} is based on the classical idea of Stein's Unbiased Risk Estimator (SURE) \citep{stein1981estimation}. SURE gives an unbiased estimate of MSE, however, it does not require ground truth data for tuning parameters of denoising algorithm thus circumventing the main hurdle for deep learning based denoisers that require clean ground truth for training. 

\begin{figure}[h]
\centering
\includegraphics[width=14cm, height=3.8cm]{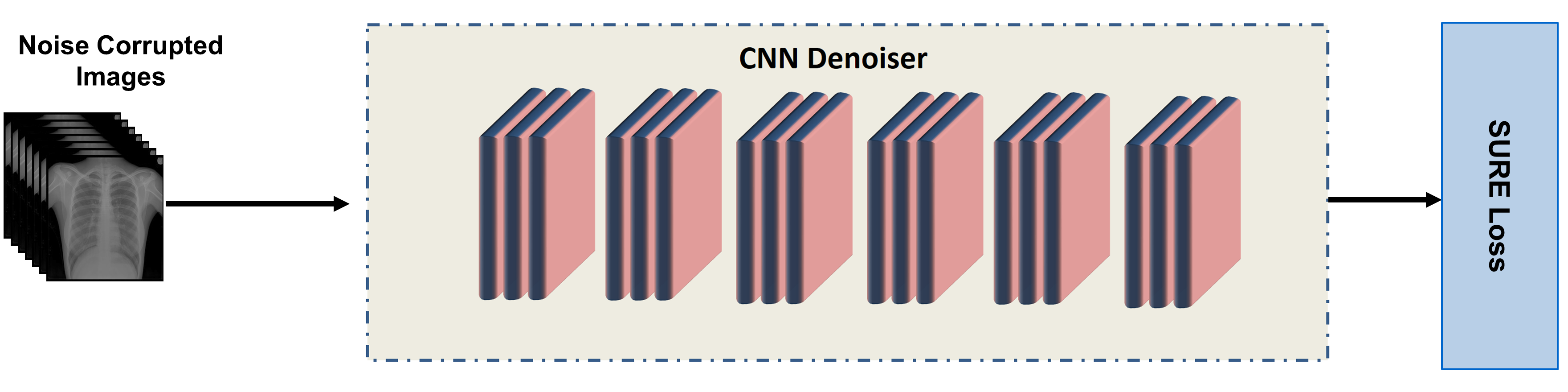}
\caption{We incorporate SURE loss (\ref{eq:SURE}) to train a convolutional neural network for X-ray image denoising. SURE loss allows us to train our network using only corrupted images without requiring ground truth clean images.}
\label{model_diag}
\end{figure}

\section{Methodology}
We consider recovering true X-ray image $\boldsymbol{x} \in \mathbb{R}^K$ from its noisy measurements of the form   
\begin{equation}
\boldsymbol{y} = \boldsymbol{x} + \boldsymbol{w},
\end{equation}
where $\boldsymbol{y} \in \mathbb{R}^K$ is noise corrupted image, $\boldsymbol{w} \in \mathbb{R}^K$ denotes independent and identically distributed Gaussian noise i.e.  $\boldsymbol{w} \sim \mathcal{N}(0,\sigma^2 \boldsymbol{I}) $ where $\boldsymbol{I}$ is identity matrix and $\sigma$ is standard deviation that is assumed to be known. We are interested in a weakly differentiable function $f_\theta(\cdot)$ parametrized by $\theta$ that maps noisy X-ray images $\boldsymbol{y}$ to clean ones $\boldsymbol{x}$. We model $f_\theta(\cdot)$ by a convolutional neural network (CNN) where $\theta$ are weights of this network. CNN based denoising methods are typically trained by taking a representative set of clean ground truth images $\boldsymbol{x}_1, \boldsymbol{x}_2,...,\boldsymbol{x}_L$ along with corresponding set of noise corrupted observations $\boldsymbol{y}_1, \boldsymbol{y}_2,...,\boldsymbol{y}_L$. The network then learns the mapping $f_\theta : \boldsymbol{y} \rightarrow \boldsymbol{x}$ from noisy observations  to clean images by minimizing a supervised loss function; typically mean squared error (MSE). MSE minimizes the error between true images and  the network output as follows:
\begin{equation} \label{eq:MSE}
\underset{\theta}{\text{min}} \quad
\sum_{\ell=1}^{L} \frac{1}{K} \Vert \boldsymbol{x}_{\ell} - f_\theta({\boldsymbol{y}_{\ell}}) \Vert^2. 
\end{equation}

Note the dependence of MSE on ground truth clean images $\boldsymbol{x}$. Instead of minimizing MSE, we employ SURE loss that optimizes neural network parameters $\boldsymbol{\theta}$ by minimizing 
\begin{equation} \label{eq:SURE}
\sum_{\ell=1}^{L} \frac{1}{K} \Vert \boldsymbol{y}_{\ell} - f_\theta({\boldsymbol{y}_{\ell}}) \Vert^2 - \sigma^2 + \frac{2 \sigma^2}{K} \text{div}_{\boldsymbol{y}_\ell}{(f_\theta(\boldsymbol{y}_\ell))},
\end{equation}

where $\text{div}(.)$ denotes divergence and is defined as
\begin{equation}
\text{div}_{\boldsymbol{y}}{(f_\theta(\boldsymbol{y}))} = \sum_{k=1}^{K} \frac{\partial f_{\theta k} (\boldsymbol{y})}{\partial y_k}
\end{equation}
The first term in (\ref{eq:SURE}), $\frac{1}{K} \Vert \boldsymbol{y}_{\ell} - f_\theta({\boldsymbol{y}_{\ell}}) \Vert^2$ minimizes the error between observations $\boldsymbol{y}$ and corresponding estimates at network output $f_\theta({\boldsymbol{y}_{\ell}})$. The second term $\frac{2 \sigma^2}{K} \text{div}_{\boldsymbol{y}_\ell}{(f_\theta(\boldsymbol{y}_\ell))}$ penalizes neural network based denoiser for varying as its input image is changed.
Calculating divergence of the denoiser is a central challenge for SURE based estimators. We estimate divergence via fast Monte Carlo approximation, see \citep{ramani2008monte} for details. In short, instead of utilizing a supervised loss of MSE in (\ref{eq:MSE}), we optimize network weights $\boldsymbol{\theta}$ in an unsupervised manner using (\ref{eq:SURE}), that does not require ground truth; see Figure \ref{model_diag}. We leverage the auto-differentiation function of Tensorflow \citep{abadi2016tensorflow} to calculate the gradient of the SURE base loss function, that is hard to compute otherwise.



\section{Experiments}
\begin{figure}[t]
\centering 
\includegraphics[scale=0.53]{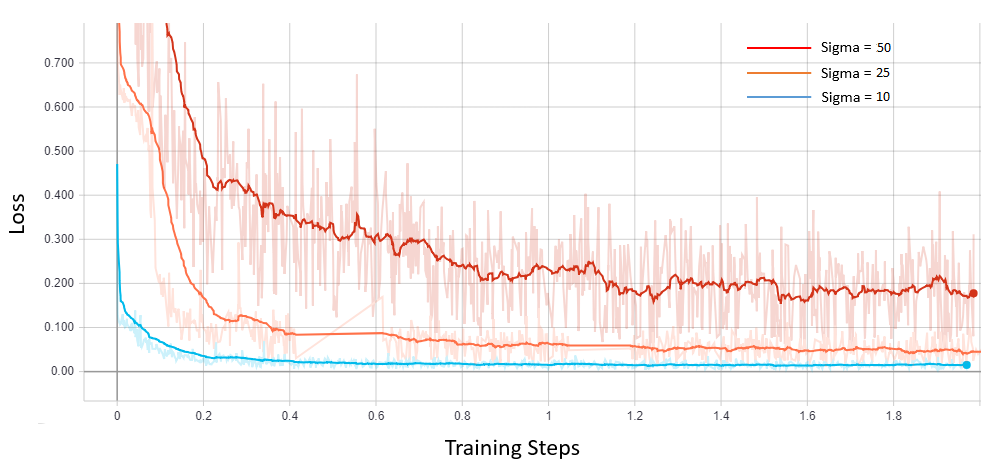}
\caption{Figure shows SURE loss as a function of number of steps of training at different noise levels. Note that higher levels of noise make it hard for the network to converge.}
\label{loss_curve}
\end{figure}

To evaluate the proposed denoising approach, we use Indiana University's Chest X-ray database \citep{IndianaXrays}. The database consists of $7470$ chest X-ray images of varying sizes, out of which we select $500$ images for training due to the scarcity of computational resources. Training images are re-scaled, cropped, and flipped, to form a set of $789,760$ overlapping patches each of size $40 \times 40$. We use an end-to-end trainable denoising convolutional neural network (DnCNN) \citep{zhang2017beyond} that have recently shown promising denoising results. DnCNN consists of 16 sequential $3 \times 3$ convolutional layers with residual connections. Training was conducted on batches of size 64 using Adam optimizer for 50 epochs with learning rate set to $10^{-4}$ which was reduced to $10^{-5}$ after 25 epochs. DnCNN was trained using SURE loss, without any ground truth clean data. We perform experiments for three different additive Gaussian noise levels having standard deviations of $10$, $25$ and $50$; see Figure \ref{loss_curve} for SURE training loss curve for each noise level. The network easily converges for low noise while higher noise levels make convergence harder. For a benchmark, we also trained DnCNN using MSE loss of (\ref{eq:MSE}) and compare its performance with the SURE approach.

\begin{figure}[h]
\centering     
\raisebox{0.1in}{\rotatebox[origin=t]{90}{\small \hspace{8em} Indiana University \citep{IndianaXrays}}} \hspace{-0.25em}
\subfigure{\label{fig:NLM_Original}\includegraphics[width=31mm,height=39mm]{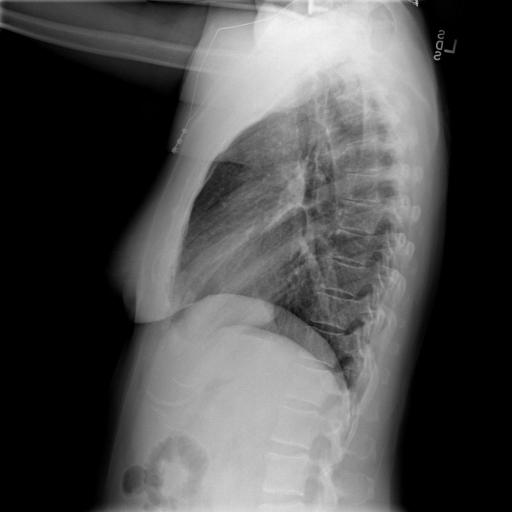}}
\subfigure{\label{fig:NLM_NOISY_25}\includegraphics[width=31mm,height=39mm]{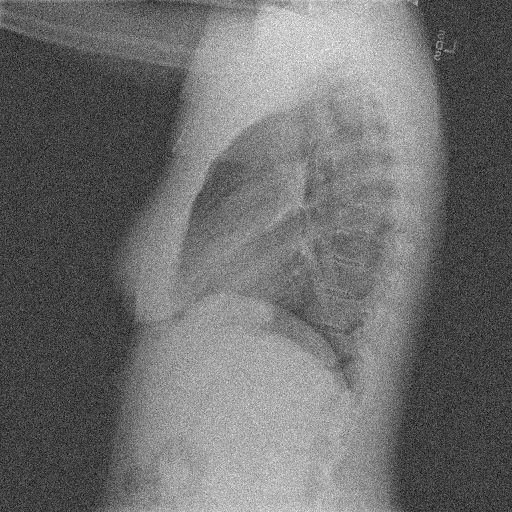}}
\subfigure{\label{fig:NLM_SURE_CLEAN_25}\includegraphics[width=31mm,height=39mm]{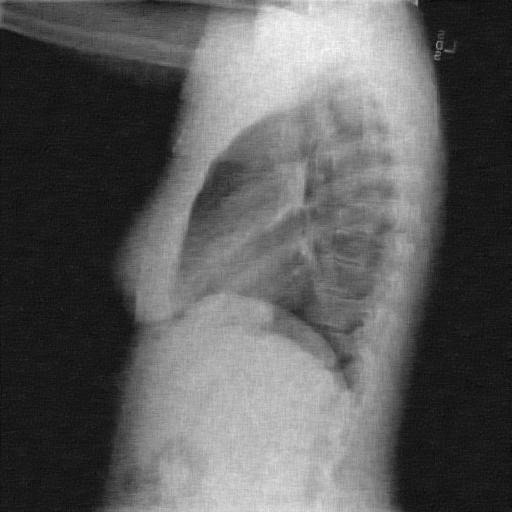}}
\subfigure{\label{fig:NLM_MSE_CLEAN_25}\includegraphics[width=31mm,height=39mm]{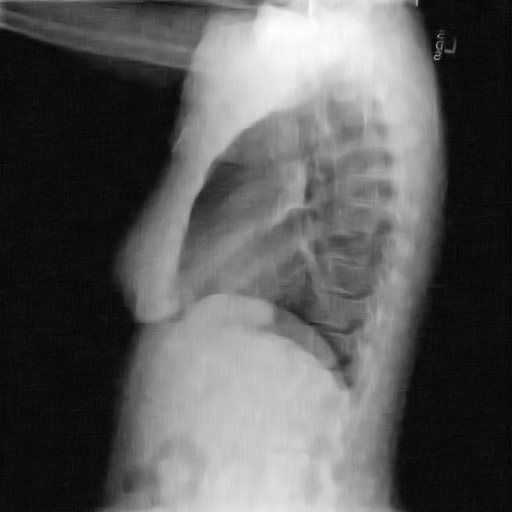}}   \\[-6.8em] 
\setcounter{subfigure}{0}
\raisebox{0.1in}{\rotatebox[origin=t]{90}{\small \hspace{8em} Chest X-Ray \citep{wang2017chestx}}} \hspace{-0.25em}
\subfigure[Original]{\label{fig:CHEST_Original}\includegraphics[width=31mm,height=39mm]{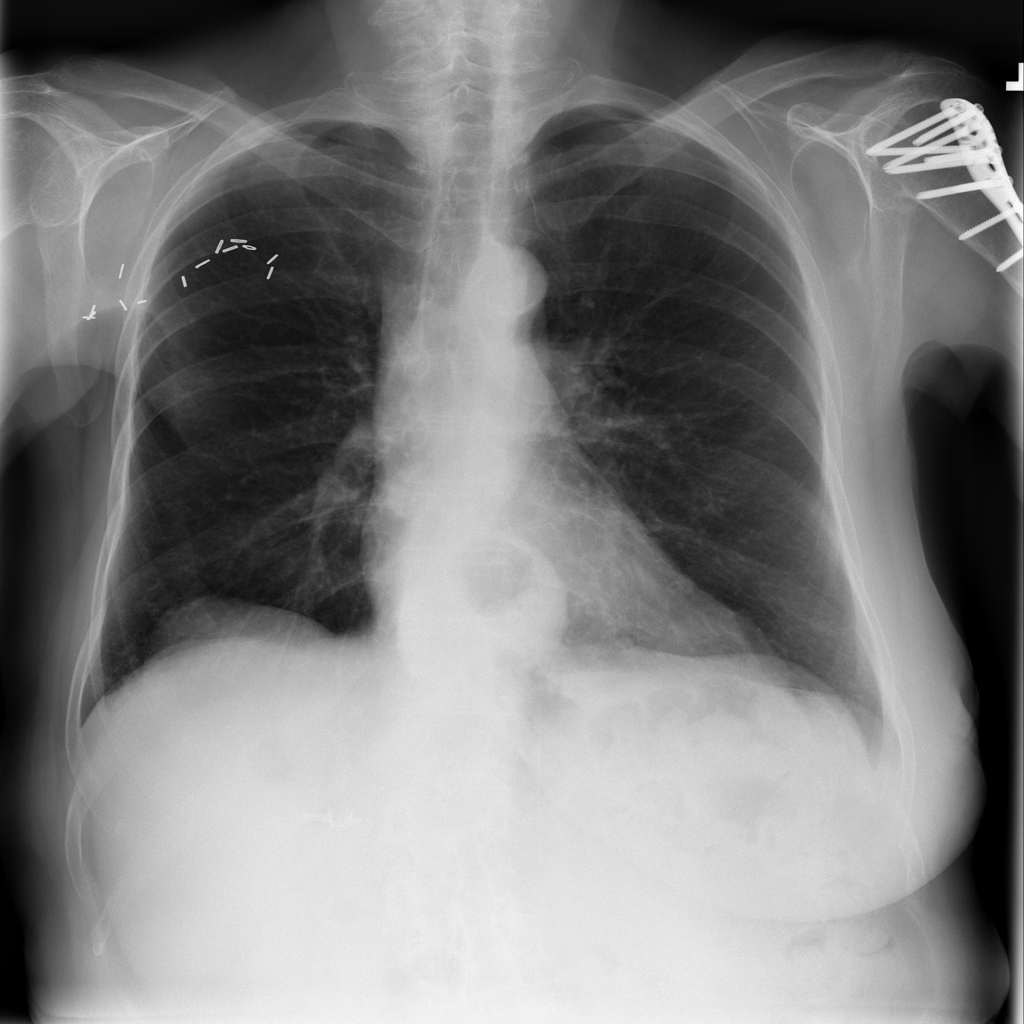}}
\subfigure[Noisy]{\label{fig:CHEST_NOISY_25}\includegraphics[width=31mm,height=39mm]{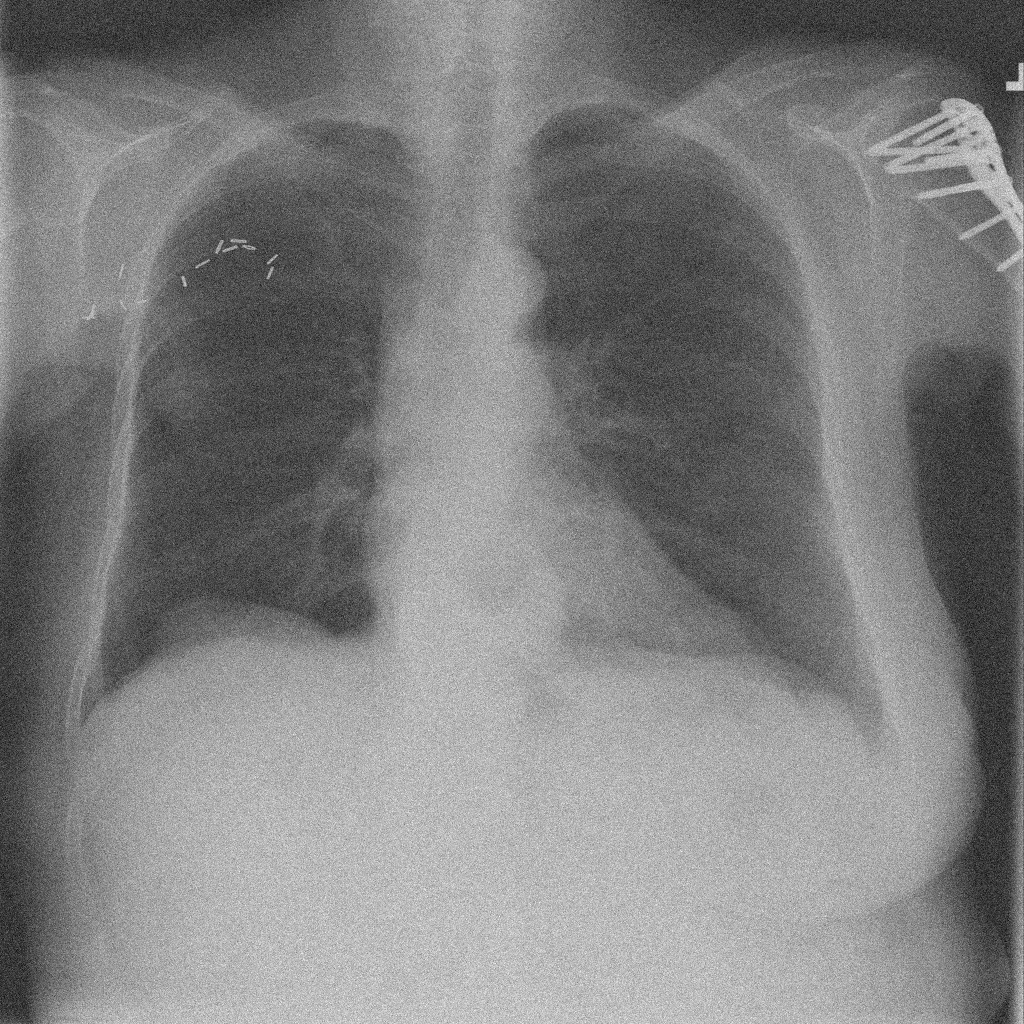}}
\subfigure[SURE]{\label{fig:CHEST_SURE_CLEAN_25}\includegraphics[width=31mm,height=39mm]{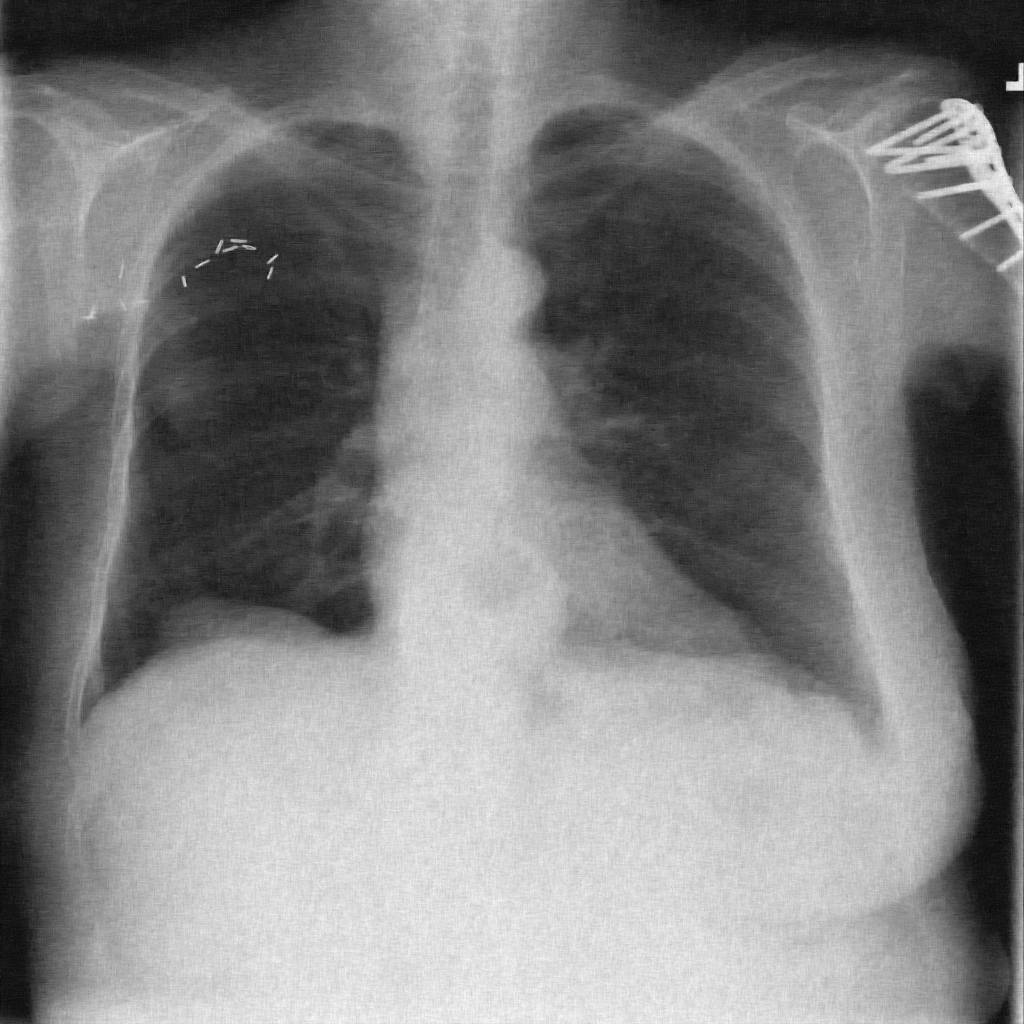}}
\subfigure[\small MSE]{\label{fig:CHEST_MSE_CLEAN_25}\includegraphics[width=31mm,height=39mm]{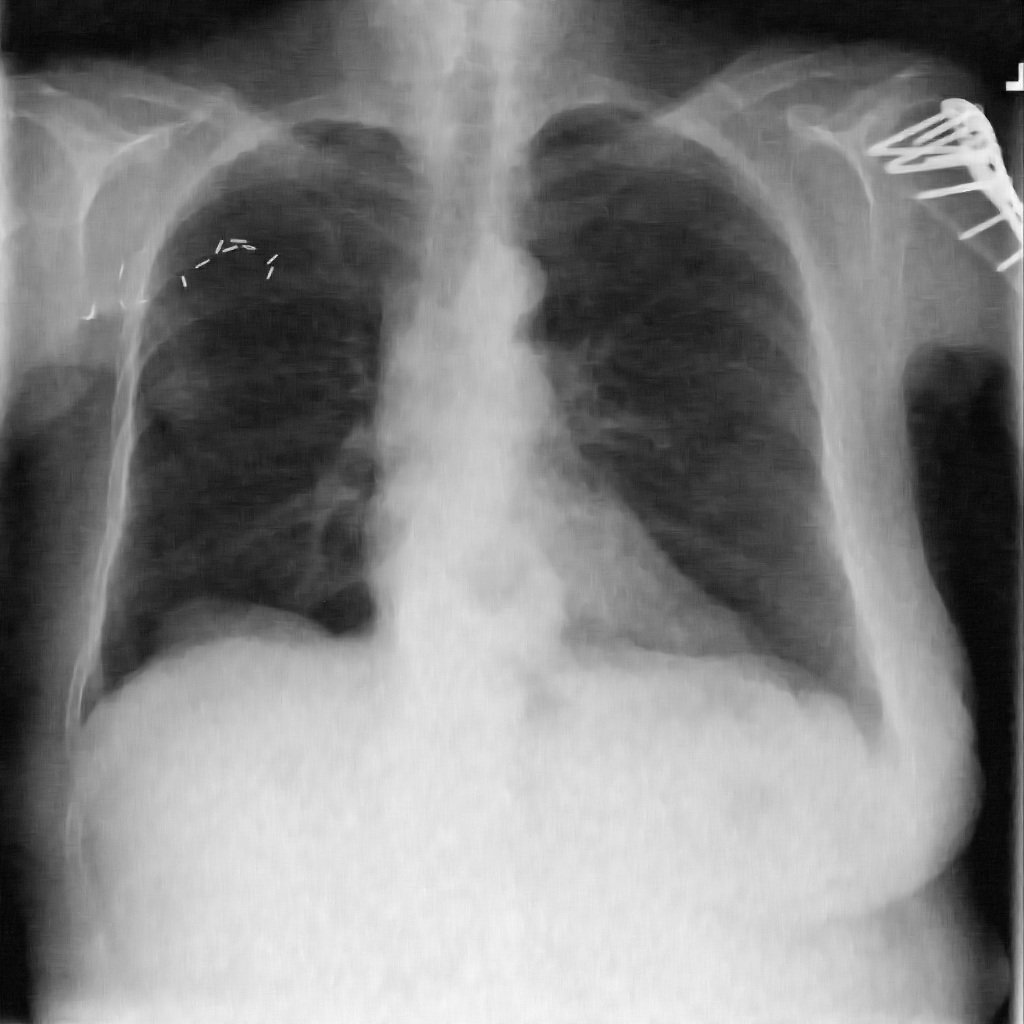}}
\caption{Denoising results for Indiana University \citep{IndianaXrays} (first row) and Chest X-Ray Dataset \citep{wang2017chestx} (second row) for Gaussian noise with standard deviation of $25$. DnCNN trained on SURE loss is able to remove noise (c) from noisy images (b), without requiring any ground truth training data. Results are comparable against supervised MSE as shown in (d), which uses ground truth images for training DnCNN.}
\label{results}
\end{figure}

For evaluation, we randomly select $10$ images from the test set of Indiana University X-Ray dataset. To quantitatively evaluate the performance of proposed SURE based approach, we use two widely used performance metrics, Peak Signal to Noise Ratio (PSNR) and Structural Similarity Index Measure (SSIM) \citep{wang2004image}. PSNR of reconstructed image $\hat{\boldsymbol{x}}$ from true image $\boldsymbol{x}$ is defined as $10 \log_{10} \frac{255^2}{\Vert \hat{\boldsymbol{x}} - \boldsymbol{x}  \Vert^2}$ for image pixels in the range of 0 and 255. On the other hand, SSIM measure perceived similarity between reconstructed and true image.
In addition to Indiana University dataset, we also use images from famous Chest X-Rays dataset \citep{wang2017chestx} for testing as well. Table \ref{performance1} shows quantitative results for both datasets at different noise levels. Figure \ref{results} shows visual results for both datasets for Gaussian noise having a standard deviation of 25. Quantitative and qualitative results show that model trained on Indiana University dataset also has very compelling results on Chest X-Ray data. This demonstrates the generalizability of SURE based approach to other datasets of similar modalities. Table \ref{comparison1} shows quantitative comparison between DnCNN trained using SURE loss and MSE loss. Note that although SURE does not require any clean ground truth images, it's performance is still comparable to DnCNN trained via supervised MSE loss, which requires ground truth images during training.   

\begin{table}[h]
\centering
\caption{Table shows average quantitative results in terms of PSNR, SSIM, and MSE of SURE based DnCNN denoiser for Indiana University X-Rays dataset \citep{IndianaXrays} and Chest X-Ray dataset \citep{wang2017chestx}. The SURE based model shows promising results, even though it does not require any clean ground truth images for training DnCNN. Interestingly DnCNN model trained only on Indiana University X-Rays dataset perform well on Chest X-Ray dataset.}
\scalebox{1}{
\begin{tabular}{|c|c|c|c|c|c|}
\hline
\multicolumn{1}{|c|}{\textbf{Dataset}} & \textbf{\begin{tabular}[c]{@{}c@{}}Noise \\ Std.\end{tabular}} & \textbf{\begin{tabular}[c]{@{}c@{}}Average\\ PSNR\end{tabular}} & \textbf{\begin{tabular}[c]{@{}c@{}}Average\\ SSIM\end{tabular}} & \textbf{\begin{tabular}[c]{@{}c@{}}Average\\ MSE\end{tabular}} & \textbf{\begin{tabular}[c]{@{}c@{}}Average \\ Time\end{tabular}} \\ \hline
\multirow{3}{*}{\begin{tabular}[c]{@{}l@{}}Indiana University \end{tabular}} & 10 & 37.50 & 0.959 & 9.12 & 0.06 \\ \cline{2-6} 
 & 25 & 32.00 & 0.881 & 13.15 & 0.06 \\ \cline{2-6} 
 & 50 & 27.78 & 0.763 & 21.50 & 0.06 \\ \hline
\multirow{3}{*}{Chest X-Rays} & 10 & 37.55 & 0.962 & 13.57 & 0.19 \\ \cline{2-6} 
 & 25 & 31.85 & 0.877 & 26.14 & 0.19 \\ \cline{2-6} 
 & 50 & 28.88 & 0.774 & 45.43 & 0.19 \\ \hline
\end{tabular}}
\label{performance1}
\end{table}

\begin{table}[!ht]
\centering
\caption{Quantitative denoising results for DnCNN trained using SURE loss and MSE loss for Gaussian noise with standard deviation of 25 and 50. Average values of PSNR, SSIM, and MSE are calculated for 10 images from test set of Indiana University X-ray dataset \citep{IndianaXrays} . }
\scalebox{1}{
\begin{tabular}{|c|c|c|c|c|}
\hline
\textbf{Method} & \multicolumn{1}{l|}{\textbf{\begin{tabular}[c]{@{}c@{}}Noise\\ Std.\end{tabular}}} & \multicolumn{1}{l|}{\textbf{\begin{tabular}[c]{@{}c@{}}Average \\ PSNR\end{tabular}}} & \multicolumn{1}{l|}{\textbf{\begin{tabular}[c]{@{}c@{}}Average\\ SSIM\end{tabular}}} & \multicolumn{1}{l|}{\textbf{\begin{tabular}[c]{@{}c@{}}Average \\ MSE\end{tabular}}} \\ \hline
\multirow{2}{*}{DnCNN-SURE} & 25 & 32.00 & 0.88 & 13.15 \\ \cline{2-5} 
 & 50 & 27.78 & 0.76 & 21.50 \\ \hline
\multirow{2}{*}{DnCNN-MSE} & 25 & 35.39 & 0.95 & 08.96 \\ \cline{2-5} 
 & 50 & 29.61 & 0.82 & 17.32 \\ \hline
\end{tabular}}
\label{comparison1}
\end{table}

\section{Discussion and Future Direction} \label{sec:discussion}
The main contribution of this work is to demonstrate the effectiveness of SURE for unsupervised X-ray denoising. Not only are we able to remove additive noise from X-ray images but also preserve the fine structure in X-ray scans. 
Our work assumes that true image is corrupted by Gaussian noise with known variance. In our future work, we will extend this SURE based approach to Poisson noise that is more relevant for X-ray imaging, especially in low dose regime. For this, we can use transforms to first convert Poisson noise to Gaussian noise and then use Gaussian noise removal methods. This is because algorithms proposed for Gaussian noise fails to give plausible results on Poisson noise. 
\section*{Acknowledgement}
We gratefully acknowledge the support of the NVIDIA Corporation for the donation of NVIDIA TITAN Xp GPU for our research.

\bibliographystyle{plain}
\bibliography{ref} 



\end{document}